\def\year{2021}\relax
\documentclass[letterpaper]{article} 
\usepackage{aaai21}  
\usepackage{times}  
\usepackage{helvet} 
\usepackage{courier}  
\usepackage[hyphens]{url}  
\usepackage{graphicx} 
\usepackage{booktabs}
\usepackage{natbib}  
\usepackage{caption} 
\usepackage[normalem]{ulem}
\useunder{\uline}{\ul}{}
\title{Negative Associations in Word Embeddings Predict anti-Black Bias Across Regions--but only via Name Frequency}
\author{
  Austin van Loon, \textsuperscript{\rm 1} 
  Salvatore Giorgi, \textsuperscript{\rm 2} 
  Robb Willer, \textsuperscript{\rm 1}
  Johannes Eichstaedt \textsuperscript{\rm 3} \\
  }
  
\affiliations{
    \textsuperscript{\rm 1}Stanford University, Department of Sociology\\
    \textsuperscript{\rm 2}University of Pennsylvania, Department of Computer and Information Sciences\\
    \textsuperscript{\rm 2}Stanford University, Department of Psychology\\
    \{avanloon,willer,eichstaedt\}@stanford.edu, sgiorgi@sas.upenn.edu

}

\begin{document}

\maketitle

\begin{abstract}
The word embedding association test (WEAT) is an important method for measuring linguistic biases against social groups such as ethnic minorities in large text corpora. It does so by comparing the semantic relatedness of words prototypical of the groups (e.g., names unique to those groups) and attribute words (e.g., `pleasant' and `unpleasant' words). We show that anti-black WEAT estimates from geo-tagged social media data at the level of metropolitan statistical areas strongly correlate with several measures of racial animus—even when controlling for sociodemographic covariates. However, we also show that every one of these correlations is explained by a third variable: the frequency of Black names in the underlying corpora relative to White names. This occurs because word embeddings tend to group positive (negative) words and frequent (rare) words together in the estimated semantic space. 
As the frequency of Black names on social media is strongly correlated with Black Americans' prevalence in the population, this results in spurious anti-Black WEAT estimates wherever few Black Americans live.
This suggests that research using the WEAT to measure bias should consider term frequency, and also demonstrates the potential consequences of using black-box models like word embeddings to study human cognition and behavior.
\end{abstract}

\noindent One of the most important innovations in the last decade of natural language processing has been the development of word embeddings algorithms \cite{mikolov2013efficient,pennington2014glove}. These tools leverage statistical patterns in large text corpora to build representations of words' meanings according to the distributional hypothesis--that words deployed in similar contexts have similar meanings. There are now several different varieties of such models, and they are widely used by social scientists, computational linguists, and practitioners~\cite{vanloonforth,nelson2021leveraging,arseniev2020machine,kozlowski2019geometry}.

These models can be used to measure the ``bias'' present in a particular text corpus. The most prominent method for doing so, the word embeddings association test (WEAT), compares the semantic relatedness of different sets of words using the cosine similarity of their vector representations to detect potentially subtle ways in which one set of targets is talked about differently than another \cite{caliskan2017semantics}. For instance, the WEAT might test whether flowers are portrayed in a more positive manner than insects in the Google News corpus, or whether women are discussed as being more related to the home than men are on Wikipedia. This approach has had a large impact on computational social science and computational linguistics, and continues to be popular.

Here, we share a cautionary tale about how the tendency of word embeddings to encode all kinds of unexpected or unintended information can lead estimates of semantic relatedness---and thus the WEAT---astray. Using a geo-tagged data set of general population US Twitter data, 
we show that estimates of anti-black bias using the WEAT at the metropolitan statistical area (MSA) level significantly predicts various measures of prejudice against African Americans in those same communities. This linguistic bias measure, for instance, predicts the average black-white Implicit Association Task (IAT) score from Project Implicit \cite{xu2014psychology} (an experimental measure of bias) originating from that MSA and the level of black-white residential segregation in the MSA. We further show that these relationships are robust to wide set of standard statistical controls. 

We go on to show, however, that each of these relationships are in fact spurious, explained away by a single omitted variable---the relative frequency of names unique to ethnic minorities in the underlying corpora. Word embeddings have a tendency to conflate term frequency and positivity, making the anti-black linguistic bias detected a complete methodological artifact. This name frequency is, however, strongly correlated with the proportion of the population that is black, which is itself strongly correlated with many of these outcomes. The result is that the WEAT \textit{appears} to be a highly sophisticated way for detecting linguistic bias, but it is in fact just a noisy proxy measuring how rare Black names are in the data. 

\textbf{Contributions.} We show empirically that a popular measure of the linguistic bias in word embeddings, the WEAT, has a tendency to conflate the frequency of category words with their positivity. In settings such as ours where the frequency of category words is correlated with meaningful information, this results in misleading omitted variable bias.

\section{Related Work}

The WEAT was introduced by \citeauthor{caliskan2017semantics} (\citeyear{caliskan2017semantics}). It takes as input a trained word embedding model, two sets of ``category words'' $A$ and $B$, and two sets of ``attribute words'' $X$ and $Y$. Where $cos(\overrightarrow{i},\overrightarrow{j})$ is the cosine similarity between the vectors assigned to words $i$ and $j$ by the trained word embedding model, it outputs a single measure of how `biased' the set of word embeddings are, measured as:

\smallskip

\begin{center}
    $S_{A,B,X,Y} = \Sigma_{x \in X} s_{x, A, B} - \Sigma_{y \in Y} s_{y, A, B}$
\end{center}

\smallskip

\begin{center}
    $s_{w, A, B} = mean_{a \in A}cos(\overrightarrow{w},\overrightarrow{a}) - mean_{b \in B}cos(\overrightarrow{w},\overrightarrow{b})$
\end{center}
\smallskip

So, for instance, $A$ and $B$ might be uniquely European-American and African-American names respectively, $X$ and $Y$ might be `pleasant' and `unpleasant' words respectively (for a review of how these attribute words are chosen, see \citeauthor{antoniak2021bad} [\citeyear{antoniak2021bad}]), and the word embedding model might be a Word2Vec model \cite{mikolov2013efficient} trained on a corpus of interest. In this case, $S_{A, B, X, Y}$ is the difference between how much more \textbf{pleasant words} are associated with white names than black names ($\Sigma_{x \in X} s_{x, A, B}$) and how much more \textbf{unpleasant words} are associated with white names than black names ($\Sigma_{y \in Y} s_{y, A, B}$). If pleasant words were found to be more associated with white names while unpleasant words were found to be more associated with black names, the overall measure would be positive, indicating anti-black bias in the underlying text corpus the word embeddings were trained on. 

This deceptively simple measure has become an integral part of the computational linguistics toolkit. Other high-profile papers such as Garg et al. (\citeyear{garg2018word}) and Lewis and Lupyan (\citeyear{lewis2020gender}) have used the WEAT to study cultural biases across time and place. Importantly, the method is now being used to evaluate the political biases of websites \cite{knoche2019identifying}, detect the purposeful spread of misinformation on social media by state-sponsored actors \cite{toney2021automatically}, uncover biases present and proliferated through popular song lyrics \cite{barman2019decoding}, and even to measure how much gender bias US judges display in their judicial opinions \cite{ash2021measuring}. 

However, there are well-known issues with word embeddings in general and the WEAT specifically that should make us skeptical of this proliferation. \citeauthor{silva2021towards} (\citeyear{silva2021towards}), for instance, find that (at least when using contextualized embedding models) WEAT estimates poorly predict bias estimated by other measures and is even internally inconsistent. \citeauthor{goldfarb2020intrinsic} (\citeyear{goldfarb2020intrinsic}) find that estimates of the bias present in word embeddings (such as those produced by the WEAT) do not meaningfully correlate with downstream biases of applications using those embeddings. 
Finally, terms in the semantic space estimated by word embeddings tend to cluster on non-intuitive dimensions such as term frequency \cite{arora2017simple,mu2017all,gong2018frage}.

We focus in on this final issue--that a term's frequency in a corpus shapes its estimated vector representation in a word embedding. Human language exhibits a clear ``linguistic positivity bias'', where positive words are used more frequently than negative words \cite{dodds2015human}. In theory, this might result in rare words being on average closer to negative words than positive words and frequent words being on average closer to positive words than negative words. \citeauthor{wolfe2021low} (\citeyear{wolfe2021low}), consistent with this reasoning, find that the degree of linguistic bias, estimated using the WEAT, towards names unique to ethnic minorities is highly correlated with the frequency with which they appear in the underlying corpus.



\section{Data and Methods}

\subsection{Geo-Tagged Twitter Data}
A random 10\% sample of the Twitter stream (i.e., the ``Garden Hose") was collected between January 2010 and May 2014, after which the data was reduced to a 1\% sample for the remainder of 2014~\cite{preotiuc2012trendminer}. Each tweet was mapped to a MSA (by first mapping to a U.S. county which is then trivially mapped to a MSA). If latitude/longitude information is available, then a tweet can trivially be mapped to a US county. If latitude/longitude data is not available for a given tweet, then location information is extracted from the self-reported User Location field, if available. This is a rule based mapping system designed to avoid false positives (i.e., incorrect mappings) at the expense of fewer mappings. For full details please see \citet{schwartz2013characterizing}. 

We also removed retweets and quoted tweets from the corpus, as we were interested only in the original language produced by MSA residents. 
All retweets and quoted tweets in our data contained ``RT @'' followed by the handle of the account the tweet was a retweet/quote of. Therefore, we excluded from analysis any tweet that contained ``RT @'' in its main body. However, some small number of tweets that were not retweets or quoted tweets likely contained this, and may have thus been unduly removed from the analysis.\footnote{To asses how accurate this heuristic was, we tested it on similar Twitter data for which we \textit{did} have ground-truth meta-data indicating whether a tweet was a retweet or quoting tweet. We found the heuristic to be over 99.995\% accurate in identifying retweets.} After excluding MSAs with less that 500k tweets (see below), the final data set consists of 1.12 billion tweets from 214 MSAs (out of 384 possible MSAs).  

\subsection{MSA-Level WEAT Estimates}


Our empirical strategy is to compare WEAT-based measures of the anti-black linguistic bias present in each MSA's Twitter discourse and compare that to other regional measures of racial animus. One straight-forward approach to this would be to train completely independent word embedding models on each MSA's respective Twitter data, and subject each of these models to the WEAT. One obstacle to realizing this strategy is that the volume of Twitter data produced by the residents of many MSAs over our observation period is relatively small. Even in our large data set, the median number of tweets (after removing retweets as specified above) in an MSA was 615,474---a smaller number than typically used for high-quality word embeddings. Further, variation in the number of tweets available for each MSA might introduce unwanted bias into our estimates. 

To overcome this limitation, we leveraged the approach taken in \citeauthor{van2020explaining} (\citeyear{van2020explaining}), which allows for estimating linguistic differences among (relatively) small sub-populations. In our case, it works by first randomly sampling a fixed number of tweets from every MSA, compiling them together, and training a word embedding model (in our case a Word2Vec model\footnote{CBOW model, vector size of one-hundred, minimum term count of ten, using negative sampling and an initial learning rate of 0.025}) on this stratified corpus. This model, built over 1 million tweets, is referred to as the ``baseline model'' and represents the consensual linguistic understanding among the MSAs. Then, for each MSA, a larger fixed number of tweets is sampled from that MSA and used to continue training the baseline model. Specifically, we sample 500k tweets per MSA, excluding those with less than 500k tweets. The resulting model is the ``updated model'', which learns the idiosyncratic linguistic norms of its MSA. This updated model is then what the WEAT is performed on. This is repeated five times and the WEAT estimates for each MSA are averaged to overcome stochastic variation in sampling and in training the embedding models.

We wanted our WEAT to be as similar to that performed by \citeauthor{caliskan2017semantics} (\citeyear{caliskan2017semantics}) as possible. To that end, we used the same list of African-American and European-American names and pleasant and unpleasant words as them. Since they used multiple lists from different sources, we simply took the union of these different lists (and excluded any which were not frequent enough in the corpus to be included in the baseline model). Just as in \citeauthor{caliskan2017semantics} (\citeyear{caliskan2017semantics}), a higher score on the WEAT indicates more anti-black bias.

\subsection{Other Variables}
We briefly describe each variable below, with full descriptions in the Supplemental Materials.

\paragraph{Racial Animus Measures.} \textbf{Implicit Bias} is derived through IAT experiments. It is represented by the average $d$-score (a measure of the difference in response latency when pleasant [unpleasant] words were paired with white [black] faces and vice versa) for all respondents in a given MSA. \textbf{Explicit Bias} measures feeling more warmly towards European Americans relative to African Americans (as self-reported). \textbf{Opposition to Affirmative Action} asks how respondents felt about affirmative action policies in general, which is averaged for all white respondents (on surveys). \textbf{Racial Resentment} is measured by racial resentment scale which attitudes thought to be indicative of racial animus among contemporary American whites (on surveys) \cite{kinder1996divided}. \textbf{Residential Segregation} is the proportion of one group that would need to change the location of their residence for the MSA to have no segregation. 

\paragraph{Relative Black Name Frequency.} For each MSA, we measure how often each name used as a category word in the WEAT appears in the MSA's Twitter discourse and find the proportion of all name occurrences that are uniquely black.

\paragraph{Controls} We collect the following information for each MSA to use as standard control variables: the proportion of the population living in poverty, the log of the total population count, the logged population density, the proportion of residents whose highest educational attainment was completing high school or less, and the proportion of households that live in a rural area. We also collect the proportion of residents that identify as black. Finally, we create a series of binary variables indicating in which of the nine census division each MSA resides. 

\subsection{Statistical Analysis}

Following the methods used in the social sciences, we use ordinary least squares (OLS) regression, which models the outcome as a weighted linear combination of the predictors with random, Gaussian-distributed noise and for every predictor yields a standardized coefficient $\beta$ with an associated significance. 
The simultaneous inclusion of multiple predictors allows OLS to partial out (control for) the  associations of different covariates with the outcome. 
In all models, all variables are standardized (mean centered and re-scaled to a unit standard deviation ) to ease comparison and interpretability. 


\section{Results}

\begin{table}[tbp]
\centering
\resizebox{.48\textwidth}{!}{
\begin{tabular}{lcccc}
\toprule
\multicolumn{1}{c}{{}} &  \begin{tabular}[c]{@{}c@{}}No \\ controls\end{tabular} & \begin{tabular}[c]{@{}c@{}}Standard \\ controls\end{tabular} & \begin{tabular}[c]{@{}c@{}}Percent \\ black\end{tabular}  & \begin{tabular}[c]{@{}c@{}} Relative \\black \\ name freq.\end{tabular} \\
\hline
Implicit bias & 0.23*** & 0.22** & 0.12* & 0.06  \\
Explicit bias & 0.23*** & 0.14* & 0.16* & 0.10  \\
Racial resentment & -0.19** & -0.10* & -0.09  & -0.05 \\
Opp. affirm. action & -0.23*** & -0.18** & -0.12 & -0.08 \\
Res. segregation & -0.19** & -0.12* & -0.12 & -0.12 \\
\bottomrule
\end{tabular}}
\caption{\small Associations between WEAT Estimates and Racial Animus Measures (rows), considering different controls (columns). Standard controls include: \% in poverty, log population, log population density, \% HS or less, \% rural, and census division. * p \textless 0.05; ** p \textless 0.01; *** p \textless 0.001.}
\label{table weat coeff}
\end{table}

Table \ref{table weat coeff} reports a summary of our results. Each cell displays the estimated $\beta$ coefficient corresponding to the strength of the relationship between an MSA's WEAT estimate and the outcome variable indicated by the row label. Each column corresponds to a set of covariates included in the model simultaneously. Asterisks denote levels of statistical significance.

In the first column labeled `No controls', coefficients are equivalent to the bivariate Pearson correlation between WEAT estimates and the outcome. As can be seen, the WEAT estimates significantly correlate with each of the outcome measures, and each correlation is highly significant. Note, however, that the direction of the correlations with racial resentment, opposition to affirmative action, and residential segregation are all in the opposite direction of what one might initially expect (i.e., MSAs with higher survey-measured racial resentment show lower WEAT-measured anti-black bias). 

The second column of Table \ref{table weat coeff} displays the same coefficient when controlling for an extensive list of standard controls at the MSA level, including census division dummies as well as proxies for socioeconomic status, average level of education, and rural/urban status. Even with these controls, the WEAT estimates strongly and significantly correlate with each of the five outcomes, suggesting robust relationships.

The third column summarizes the results of models that only control for percent of the MSA that identifies as black (and does not include the several covariates included under `Standard controls'). As can be seen in the table, in most cases the coefficient magnitude is reduced more when controlling for this one variable than when controlling for the several covariates included the previous column. However, the relationships between WEAT estimates and implicit bias as well as explicit bias remain statistically significant.

The fourth and final column of Table \ref{table weat coeff} shows the coefficient of the relationship between WEAT estimates and each outcome when controlling for relative black name frequency. As can be seen, none of the relationships attain statistical significance. This indicates that the WEAT estimates don't contain significantly more information regarding the outcomes than the name frequencies alone. 

\begin{figure}
    \centering
    \includegraphics[width=0.45\textwidth]{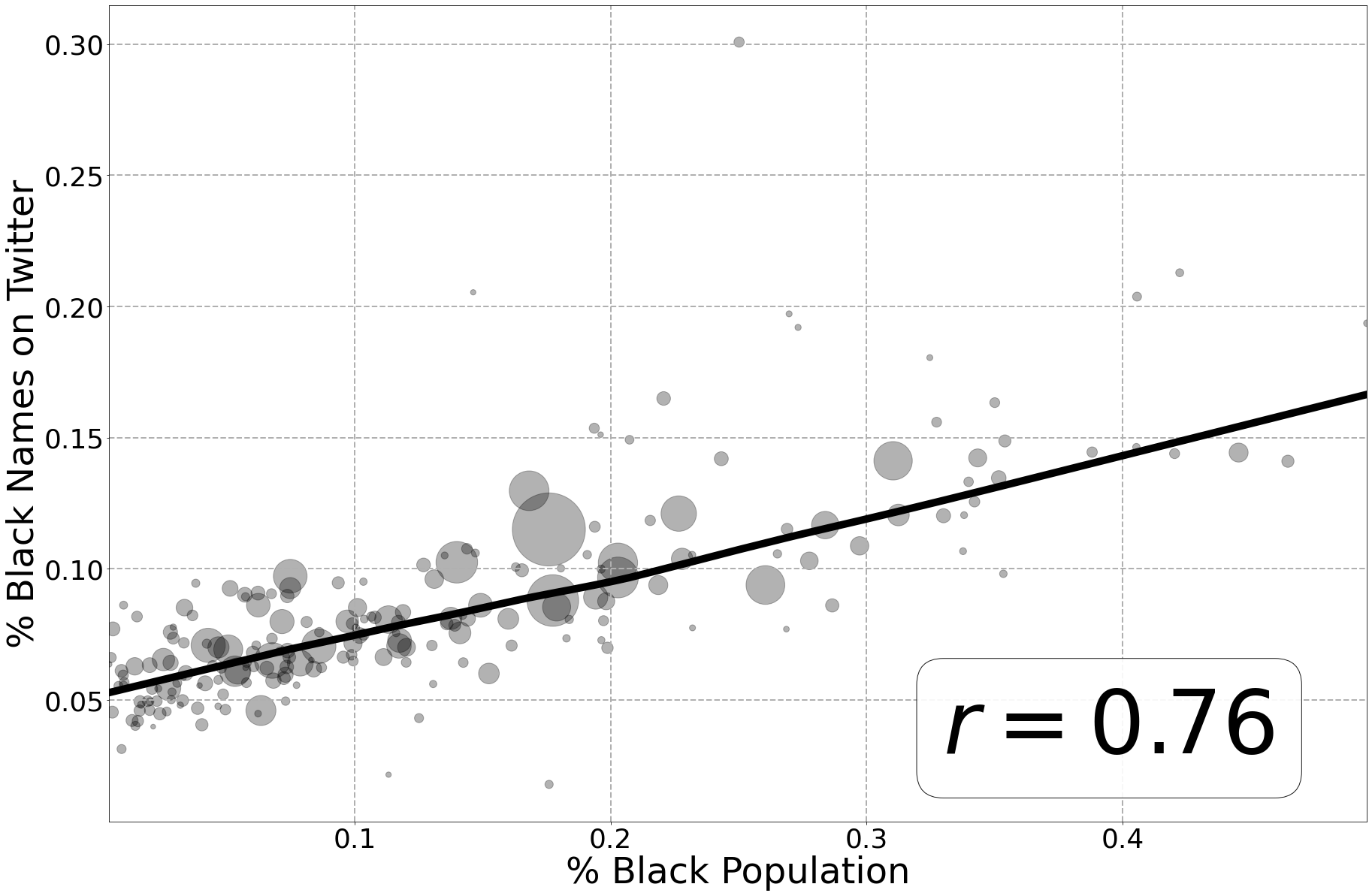}
    \caption{\small The relationship between the proportion of an MSA's populaton that is black and the proportion of names used in the WEAT that were uniquely black. Observations are linearly scaled in size by total population count. Line is LOWESS fitted to these points. 
    }
    \label{fig:lowess}
\end{figure}

To unpack the difference between columns 3 and 4, see Figure \ref{fig:lowess}, which shows that that the proportion of the population that identifies as black and the proportion of occurrences of uniquely black names in the WEAT are highly correlated (explaining over half of the variance). That is, Black names are used more in areas where Black Americans reside--and while controlling for \% Black Population partially accounts for the association between the WEAT and Racial Animus Measures, it is really the relative occurrence of Black names that fully accounts for these associations.

\section{Conclusion}

Word embeddings are an undeniably powerful tool for the study of human language and cognition. A prominent article in the \textit{American Sociological Review} has even said that they reveal the very ``geometry of culture'' \cite{kozlowski2019geometry}. However, these models are also black-boxes; they seem to provide valuable information, but due to their complexity researchers cannot easily observe \textit{how} they arrive at that information.

In this work, we showed that one potentially unintuitive aspect of word embeddings (their tendency to separate rare and common words in their estimated semantic space) can have unintended consequences for the study of human attitudes. Specifically, when estimating latent linguistic biases against social groups using the popular Word Embedding Association Test (WEAT), these models can conflate the relative frequency of words prototypical of the groups with positivity. This is especially problematic in our setting where we analyzed linguistic bias against groups with varying prevalences in the text-generating populations, which corresponds to a commensurate lack of representation in the Twitter data. In sum, this created spurious relationships between our estimates of linguistic bias and various experimental and survey-based measures of regional racial animus.

We were able to uncover this tendency by showing that these relationships vanished upon controlling for a single variable: the frequency of the words prototypical of the minority group relative to that same frequency for the majority group. While it's relieving that we were able to find a simple way to alleviate this omitted variable bias, it's worrisome that a different but fairly extensive set of controls did not sufficiently correct for it. This indicates that if other biases we don't know about are also introduced by the use of word embeddings, we might not be able to rely on standard sociodemographic controls to fully address them.


The current work relies on the widely used Word2Vec model \cite{mikolov2013efficient}; future work may extend analysis to GloVe models \cite{pennington2014glove} and contextual embeddings models such as BERT \cite{devlin2018bert}, as well as consider other measures of linguistic bias beyond the WEAT.

Our findings have important consequences for computational linguistics and computational social science. First and foremost, research using the WEAT in a way similar to how we do here should strongly consider either measuring and controlling for the relative frequency of the seed words used in the WEAT or estimating their word embeddings such that they are frequency-agnostic \cite{mu2017all,gong2018frage}. Second, other measures for assessing linguistic bias should be audited to uncover whether or not similar biases exist as in the WEAT. Finally, social scientists using word embeddings models should take heed: these models are complex and their validity has not been properly demonstrated. Careful study of these models with this goal in mind is necessary before they can truly be considered a measure of the ``geometry of culture.''




\bibliography{bib.bib}

\end{document}


\maketitle

\subsection{Variable Descriptions}

\noindent \textit{Implicit Bias}. All publicly available sessions of the race implicit association test (IAT; see \citeauthor{greenwald1998measuring} [\citeyear{greenwald1998measuring}]) were downloaded from Project Implicit ((https://osf.io/y9hiq/). All responses for which MSA information or responses for feeling thermometer measures (see below) were unavailable were excluded. Then, the average $d$-score (a measure of the difference in response latency when pleasant [unpleasant] words were paired with white [black] faces and vice versa) for all respondents in a given MSA\footnote{The number of sessions for an MSA ranged from 368 to 237576, with a mean of 15558.71 and a median of 6372. Results are substantively identical for implicit and explicit bias when excluding MSAs with fewer than 500, 1000, or 5000 sessions.} was calculated as the measure of implicit bias.

\noindent \textit{Explicit Bias}. After completing the race IAT, respondents were asked on a 10-point feeling thermometer how warmly they felt towards African Americans and European Americans. For each of the sessions used to calculate implicit bias (see above), the difference score was taken such that higher scores indicated feeling more warmly towards European Americans relative to African Americans. The average of this score for all sessions originating in each MSA was then taken. Implicit bias and explicit bias at the MSA level are highly correlated ($r = 0.89$).

\noindent \textit{Opposition to Affirmative Action}. The Cooperative Election Study (CES; formerly the Cooperative Congressional Election Study or CCES) is a yearly, nationally stratified sample survey administered in the US by YouGov (see https://cces.gov.harvard.edu/). During some years, the CES asked how respondents felt about affirmative action policies in general. We averaged responses of all white respondents\footnote{These measures are thought to be more related to racial animus for white Americans than other groups.} in each MSA, harmonizing the scales used in different years by setting them all to be between zero and one (see \citeauthor{acharya2016political} [\citeyear{acharya2016political}]). Results are substantively identical when reducing all responses to ``oppose'' or ``not oppose'' and taking a proportion.

\noindent \textit{Racial Resentment}. The symbolic racism (or racial resentment) scale is meant to measure attitudes thought to be indicative of racial animus among contemporary American whites \cite{kinder1996divided} and is widely used in psychology and political science. In various years of the CES, parts of this scale were asked of respondents. We harmonized responses across years such that the sum of responses range from zero to one, and take the average of all such indexes for the white respondents in the CES for each MSA.

\noindent \textit{Residential Segregation}. The index of dissimilarity (a measure of how unevenly two groups' residences are distributed over a defined area) for each MSA in 2010 was downloaded from the Diversity and Disparities compendium (https://s4.ad.brown.edu/projects/diversity/index.htm). This measure ranges from zero to one-hundred, and can be interpreted as the proportion of one group that would need to change the location of their residence for the MSA to have no segregation.


\noindent \textit{Relative Black Name Frequency}. For each MSA, we measure how often each name used as category words in the WEAT appears in the MSA's Twitter discourse and find the proportion of all name occurrences that are uniquely black.

\section{Scatter Plots}
\begin{figure*}[!ht]
    \centering
    \includegraphics[width=\textwidth]{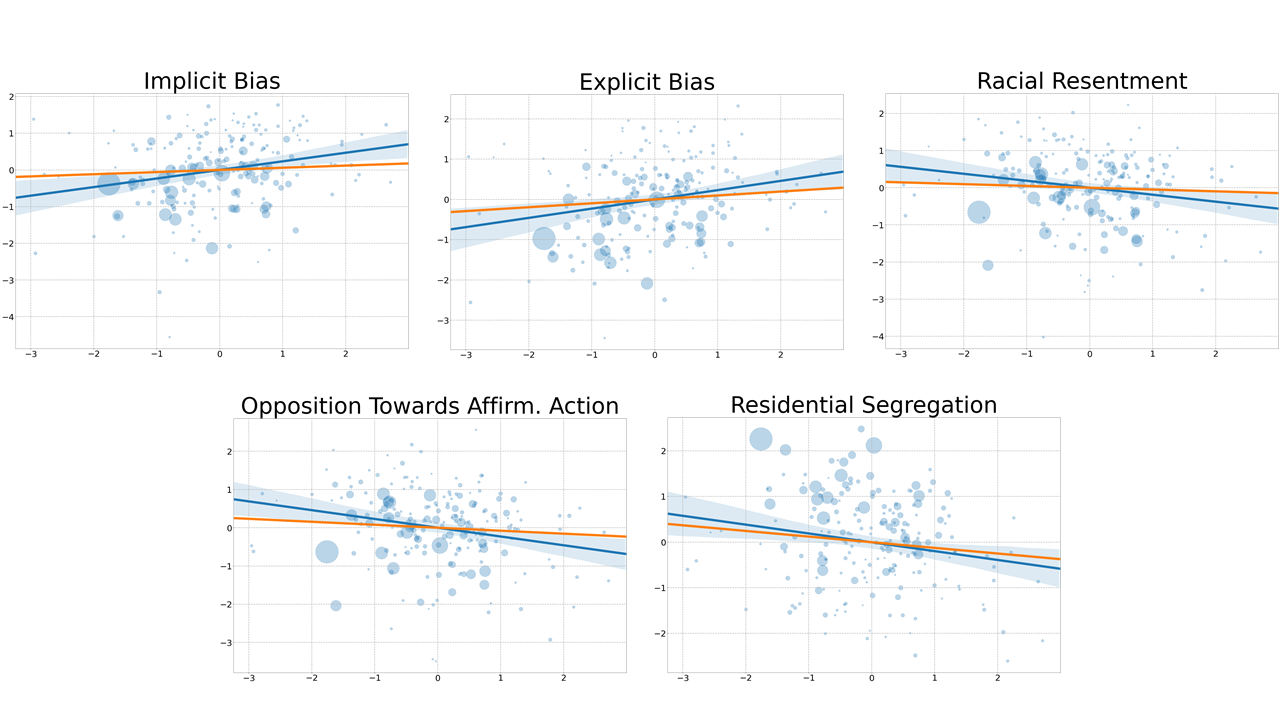}
    \caption{Scatter-plots of standardized values of MSA-level WEAT estimates (x-axis) and various outcome measures (y-axis; labeled above graph). Observations are linearly scaled in size by total population count. The blue line indicates the OLS-based line of best fit for each plot. The orange line is the best fit line after partialling out relative name frequency.}
    \label{fig:regressionPlots}
\end{figure*}

Figure \ref{fig:regressionPlots} shows the scatter plots corresponding to No controls and Relative black name frequency columns in Table 1 in the main paper.

\bibliography{bib.bib}